\providecommand{\releaseversion}{1}
\RequirePackage[OT1]{fontenc}
\documentclass[letterpaper,10pt,conference]{ieeeconf}

\IEEEoverridecommandlockouts
\usepackage{mathptmx}
\usepackage{times}

\usepackage{amsmath,amssymb}

\usepackage{graphicx}
\usepackage{overpic}
\usepackage{subcaption}
\usepackage{placeins}
\usepackage{needspace}
\usepackage{float}

\graphicspath{{figures/}}

\usepackage{booktabs,multirow,array,tabularx,textcomp}
\usepackage[table]{xcolor}
\usepackage{soul}
\usepackage{CJKutf8}

\usepackage{cite}
\makeatletter
\let\NAT@parse\undefined
\makeatother
\usepackage[numbers,sort&compress]{natbib}
\PassOptionsToPackage{hyphens}{url}
\usepackage{xurl}
\usepackage[breaklinks,colorlinks,citecolor=blue]{hyperref}

\newcommand{\setfloatblocksep}[1]{%
  \setlength{\floatsep}{#1}%
  \setlength{\textfloatsep}{#1}%
  \setlength{\dblfloatsep}{#1}%
  \setlength{\dbltextfloatsep}{#1}%
  \setlength{\intextsep}{#1}%
}
\setfloatblocksep{9pt}
\makeatletter
\renewcommand{\section}{\@startsection{section}{1}{\z@}%
  {1.5ex}{0.7ex}%
  {\normalfont\normalsize\centering\scshape}}
\renewcommand{\subsection}{\@startsection{subsection}{2}{\z@}%
  {1.0ex}{0.01pt}{\normalfont\normalsize\itshape}}
\makeatother
\DeclareRobustCommand{\matchercompass}{\textup{\textbf{\textsc{MatcherCompass}}}}
\title{\LARGE\bfseries
\matchercompass{}: A Deployment-Aware Benchmark\\
to Guide Image Matcher Selection in the Wild
}

\ifnum\releaseversion=1\relax
\author{Hyunwoo Kim$^{1}$ and Giseop Kim$^{1*}$%
\thanks{$^{1}$H. Kim and G. Kim are with the Department of Robotics and Mechatronics Engineering, DGIST, Daegu, Republic of Korea
{\tt\small [gudens0823, gsk]@dgist.ac.kr}}%
\thanks{$^{*}$Corresponding author.}}
\hypersetup{pdfauthor={Hyunwoo Kim, Giseop Kim}}
\else
\author{Anonymous Authors}
\hypersetup{pdfauthor={Anonymous Authors}}
\fi

\begin{document}

\maketitle
\thispagestyle{empty}
\pagestyle{empty}
\flushbottom

\begin{abstract}
Field robots operating across time of day and sensing modalities require accurate image
correspondences within onboard time and resource budgets. However, accuracy and
runtime reported for individual methods on a single device provide limited guidance for choosing a
matcher and its configuration on a target platform. We present \matchercompass{}, a deployment-aware
benchmark for choosing local feature matchers in field robotics. Under common input and
pose-evaluation procedures, we compare nine classical and learned matching pipelines across four
image resolutions and supported numerical precisions. Four visual conditions cover viewpoint
variation, day--night matching in visible and thermal imagery, and daytime visible--thermal
matching. We evaluate pose accuracy using the area under the
error--recall curve (AUC) at 5$^\circ$, 10$^\circ$, and 20$^\circ$, and measure runtime, GPU memory,
and energy per image pair on four GPU platforms spanning workstation and onboard computers. The
results show that changes in hardware, input resolution, and numerical precision can move a matcher across a runtime
budget boundary, altering the feasible choices. We organize the measurements into a selection guide
that returns all configurations satisfying user-specified time and resource limits, together with
their accuracy under the selected visual condition. \matchercompass{} provides measured evidence for
choosing matching pipelines that fit a robot's sensing conditions and computing hardware.

Project page: \url{https://matchercompass.github.io/}.
\end{abstract}

\section{Introduction}
\label{sec:introduction}

Deploying robots in the wild requires perception that works under the sensing conditions and
computational limits of the intended mission. These practical challenges are a sustained focus
of recent ICRA workshops, including Robots in the Wild and the Workshop on Field
Robotics~\cite{icra2025robotswild,icra2026fieldrobotics}. For visual localization and motion
estimation, feature matching must establish accurate correspondences within the robot's
processing cycle. Revisiting a place at another time of day changes image appearance; matching
visible and thermal observations additionally changes the sensing modality. Onboard time,
memory, and energy budgets constrain which matching configurations can support these tasks.

\begin{figure}[!t]
\centering
\includegraphics[width=\columnwidth]{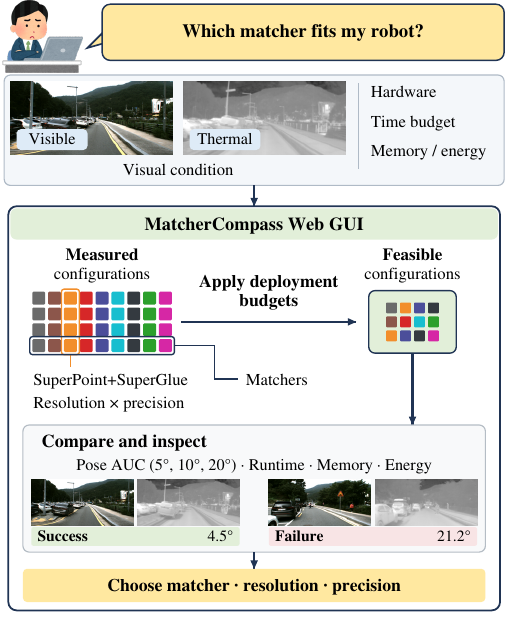}
\caption{\matchercompass{} selection workflow. Tiles are schematic and colored by matcher. Examples show pose error (success $\leq10^\circ$); images are cropped for display.}
\label{fig:selection-guide}
\label{fig:web-gui-demo}
\end{figure}

\begin{figure*}[!t]
\centering
\includegraphics[width=0.9\textwidth,keepaspectratio]{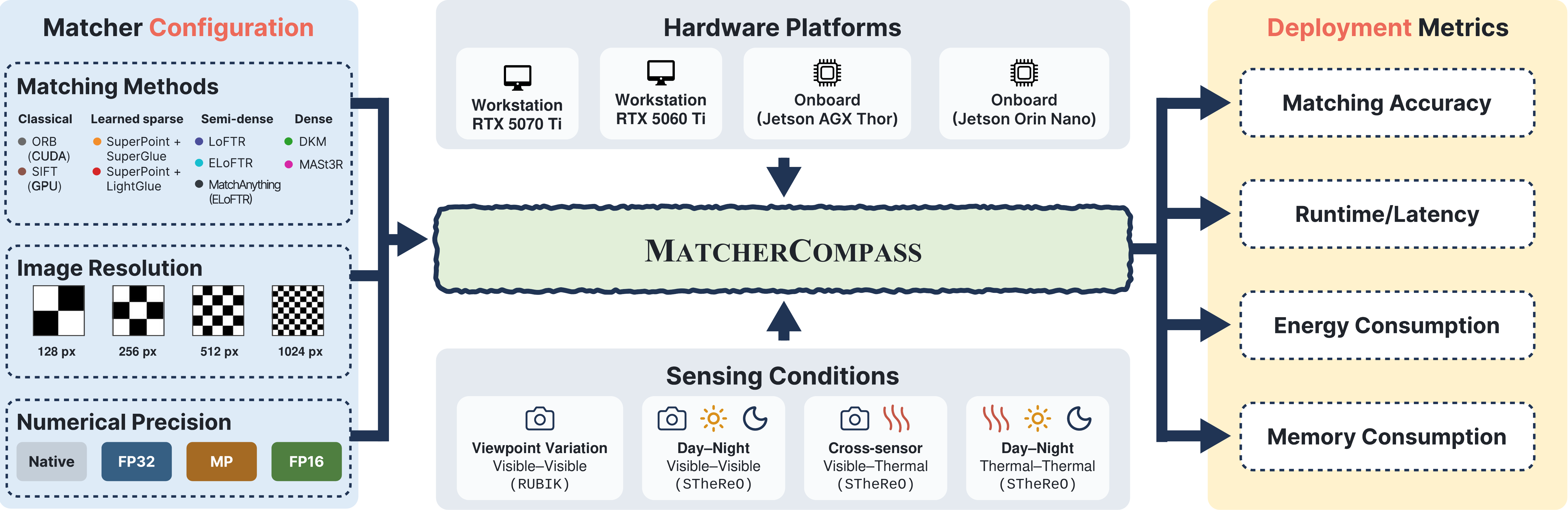}
\caption{Overview of \matchercompass{}. Accuracy and computational cost are evaluated across matchers, resolutions, and precisions under different sensing conditions and hardware platforms to support deployment-aware selection.}
\label{fig:matchercompass-flow}
\end{figure*}

Local feature matching offers classical, learned sparse, semi-dense, and dense pipelines.
Attention-based matching and geometric representations improve correspondence estimation
\cite{sarlin2020superglue,leroy2024mast3r}, while efficient designs reduce computation
\cite{wang2024eloftr}. Yet accuracy under RGB viewpoint changes does not establish suitability
for day--night or visible--thermal matching. Thermal sensing introduces another operating
condition to assess, not an assumption of improved matching accuracy. A deployment-oriented
comparison must therefore cover both geometric variation and changes in time of day and
sensor modality, while measuring the costs of the candidate configurations on the target hardware.

Joint accuracy and computational evaluation has informed hardware-dependent algorithm selection
in visual-inertial odometry~\cite{delmerico2018vio}. For feature matching, public tools such as
Image Matching WebUI allow direct matcher trials and correspondence inspection
\cite{imageMatchingWebUI}, but individual examples do not establish accuracy across the intended
conditions or compliance with onboard resource budgets. Developers need a selection guide that
connects condition-specific pose accuracy with the runtime, memory, and energy of each matcher,
resolution, and precision configuration.

We introduce \matchercompass{}, a deployment-aware benchmark for choosing image matchers
in the wild (Fig.~\ref{fig:matchercompass-flow}). Nine pipelines are evaluated across four visual
tasks and four GPU platforms, varying resolution and supported precision under common
pose-evaluation procedures. Pose AUCs at 5$^\circ$, 10$^\circ$,
and 20$^\circ$ are linked to runtime, GPU memory, and energy per pair. For a selected sensing
condition and platform, the guide returns all measured configurations meeting the specified
resource limits, with their accuracy and costs.
The interactive selection guide and benchmark results are available at
\mbox{\url{https://matchercompass.github.io/}}.

The contributions of this work are:

\begin{itemize}
\item A common evaluation of nine matching pipelines over four visual conditions, four image resolutions,
supported numerical precisions, and four GPU platforms.
\item An empirical comparison of how hardware and numerical precision change the available matcher
configurations under 20, 50, and 100 ms runtime budgets, together with the pose
accuracy, memory, and energy costs associated with each choice.
\item A selection guide that exposes all configurations satisfying those budgets, supported by recorded
evaluation settings and measurement procedures.
\end{itemize}

\section{Related Work}
\label{sec:related_work}

\subsection{Local Feature Matchers}
\label{sec:related_matchers}
Classical matchers such as SIFT~\cite{lowe2004sift} and ORB~\cite{rublee2011orb}
use hand-crafted descriptors. Learned sparse pipelines combine SuperPoint
features~\cite{detone2018superpoint} with SuperGlue's attention-based
assignment~\cite{sarlin2020superglue} or LightGlue's adaptive
computation~\cite{lindenberger2023lightglue}. Detector-free LoFTR~\cite{sun2021loftr}
and ELoFTR~\cite{wang2024eloftr} provide semi-dense correspondences, while
DKM~\cite{edstedt2023dkm} and MASt3R~\cite{leroy2024mast3r} use dense matching
representations. MatchAnything targets generalization across sensing modalities
through large-scale pre-training~\cite{He2025MatchAnything}.

These methods offer different correspondence densities, computational structures, and training
coverage. Accuracy alone does not establish which pipeline fits a field robot's hardware and
operating conditions. \matchercompass{} therefore compares complete pipelines under the same visual
conditions while varying hardware, resolution, and numerical precision.

\subsection{Benchmarking and Deployment Support}
\label{sec:related_benchmarks}
\texttt{HPatches} evaluates illumination and viewpoint changes~\cite{balntas2017hpatches}.
\texttt{RUBIK} structures relative-pose evaluation by overlap, scale, and viewpoint, and also reports
runtime~\cite{loiseau2025rubik}. \texttt{STheReO} provides stereo thermal data for odometry and mapping~\cite{yun2022sthereo}; here, its visible and thermal observations extend matcher evaluation to temporal and cross-modal conditions. Deployment additionally requires costs measured on the target hardware.

SLAMBench2 jointly evaluates accuracy, speed, memory, and power~\cite{bodin2018slambench2}.
Delmerico and Scaramuzza compare visual-inertial odometry on laptop and onboard platforms,
with interactive results for algorithm--hardware combinations~\cite{delmerico2018vio}.
STAR-Bench studies accuracy, latency, and energy for onboard satellite image registration
across sensing modalities~\cite{disalvo2026starbench}. These studies establish the value of
joint algorithm--hardware evaluation. Image Matching WebUI supports direct matcher trials
and correspondence inspection~\cite{imageMatchingWebUI}.

For field-robot matcher selection, the remaining need is to connect visual-condition-specific
accuracy to the costs of individual matcher, resolution, and precision configurations.
\matchercompass{} provides this connection and a Web GUI for finding configurations that satisfy
user-specified time and resource limits on a target platform.

\section{\matchercompass{}}
\label{sec:matchercompass}

\matchercompass{} links benchmark measurements to practical matcher selection. We first define the
configurations to compare, then the accuracy and resource metrics used to assess them, and finally
the Web GUI that helps users identify configurations meeting their deployment constraints.

\subsection{Benchmark Configurations}
The four tasks translate sensing requirements into matching evaluations
(Table~\ref{tab:visual-tasks}). \texttt{RUBIK} covers geometric variation within visible imagery.
\texttt{STheReO} complements it with day--night matching within each modality
and daytime visible--thermal matching across sensors. Together they test whether candidate configurations
remain accurate when viewpoint, time of day, or modality changes. Section~\ref{sec:experiments}
details the accuracy and resource protocols.

Each configuration specifies a matcher, input resolution, and numerical precision.
Nine pipelines at 128, 256, 512, and 1024~px (longest side), with supported Native,
FP32, MP, and FP16 modes, yield 84 configurations per task.
Table~\ref{tab:precision-support} lists implementations and precision support.

\subsection{Accuracy and Resource Metrics}
\label{sec:metrics}

Matching accuracy is evaluated using the rotation and translation-direction errors of relative poses
estimated from matched correspondences. Following SuperGlue~\cite{sarlin2020superglue}, we define
pose error as the larger of the two angular errors for each image pair and report pose area under
the curve (AUC) at 5$^\circ$, 10$^\circ$, and 20$^\circ$. AUC integrates the cumulative fraction of
pairs as a function of the error threshold up to the specified angle, normalizes by that angle, and
is reported on a 0--100 scale. We use the trapezoidal integration procedure in the public SuperGlue
evaluation code \cite{sarlin2020superglue}. All 100 pairs per task are included, with failed pose
estimates assigned infinite error.

Computational cost is evaluated through runtime (ms), GPU memory (GiB), and energy per pair (J).
Runtime is the elapsed time from a matcher call through correspondence return and GPU completion,
including preprocessing within the matcher and CPU--GPU transfers. Common image loading and
resizing, model loading, warm-up, and pose estimation are outside the timed interval. GPU memory is
defined as the peak process CUDA allocation observed during the resource measurement interval after
warm-up. Energy per pair is the integral of measured power over repeated matching divided by the
number of completed calls, without subtracting idle power.

Resource metrics are medians over four pairs per task and platform. Out-of-memory (OOM) failures
are distinguished from unsupported precision modes and other missing measurements.

\subsection{\matchercompass{} Web GUI}
\label{sec:web_gui}
The \matchercompass{} Web GUI makes these benchmark results available for deployment-aware
selection. Users specify a visual condition, target platform, and runtime and resource limits.
The interface returns all measured matcher, resolution, and precision configurations satisfying
these constraints, together with their accuracy and resource costs. For example, a 20~ms runtime
limit filters the results to configurations meeting that budget (Fig.~\ref{fig:web-gui-demo}).

\section{Experiments}
\label{sec:experiments}

\subsection{Datasets}
\label{sec:datasets}
Table~\ref{tab:visual-tasks} pairs geometric evaluation on \texttt{RUBIK} with temporal and cross-modal evaluation on \texttt{STheReO}. All configurations use the same 100 pairs per task.
\begin{table}[!htbp]
\centering
\captionsetup{justification=raggedright,singlelinecheck=false}
\caption{Two datasets and four visual matching tasks.}
\label{tab:visual-tasks}
\begingroup
\fontsize{8}{9}\selectfont
\setlength{\tabcolsep}{3pt}
\renewcommand{\arraystretch}{1.08}
\begin{tabularx}{\columnwidth}{@{}c c c >{\centering\arraybackslash}X@{}}
\toprule
Sensing & Dataset & Image pair & Variation \\
\midrule
\multirow{4}{*}[-2.4pt]{\textbf{Visible}}
& \multirow{2}{*}{\texttt{RUBIK}~\cite{loiseau2025rubik}}
& \multirow{2}{*}{Visible--visible} & Viewpoint, \\
& & & scale, overlap \\
\cmidrule[0.25pt](lr){2-4}
& \texttt{STheReO}~\cite{yun2022sthereo}
& \multirow{2}{*}{Visible--visible} & \multirow{2}{*}{Day--night} \\
& 01\,$\leftrightarrow$\,03 & & \\
\midrule[0.25pt]
\multirow{4}{*}[-2.4pt]{\textbf{Unconventional}}
& \texttt{STheReO}~\cite{yun2022sthereo}
& \multirow{2}{*}{Visible--thermal} & \multirow{2}{*}{Modality} \\
& 01 & & \\
\cmidrule[0.25pt](lr){2-4}
& \texttt{STheReO}~\cite{yun2022sthereo}
& \multirow{2}{*}{Thermal--thermal} & \multirow{2}{*}{Day--night} \\
& 01\,$\leftrightarrow$\,03 & & \\
\bottomrule
\end{tabularx}
\endgroup
\par\vspace{4pt}
\begin{minipage}{\columnwidth}
\footnotesize\raggedright
\texttt{STheReO} uses Valley01 (day, \mbox{10:53--11:00}) and Valley03
(night, \mbox{23:13--23:19}).\\
Per task: 100 pairs for accuracy; four pairs for resource measurements on each hardware platform.
Pair construction is described in Section~\ref{sec:datasets}.
\end{minipage}
\end{table}

For \texttt{RUBIK}, we manually select 11 of the benchmark's 33 difficulty levels and allocate pairs
proportionally, sampling at regular intervals in parallax order within each difficulty level~\cite{loiseau2025rubik}.

We use \texttt{STheReO} because its visible and thermal observations, repeated traversals, calibration, and reference trajectories support all three complementary tasks~\cite{yun2022sthereo}. We use \mbox{Valley01} and \mbox{Valley03} (morning and evening in the downloads), hereafter called day and night, respectively. We select 100 shared locations with median pairwise
parallax of at least 1$^\circ$ in all three tasks to avoid insufficient geometry for translation estimation.
Day--night pairing uses pose proximity; a 2~s offset provides a temporal baseline
for both day--night and daytime visible--thermal matching, and thermal frames are associated by timestamp.\footnote{Pose association uses horizontal distance within 1~m and yaw within 5$^\circ$; thermal timestamps must agree within 50~ms.}
For accuracy evaluation, visible images are demosaiced from BGGR Bayer data into RGB.
Thermal intensities use common clipping bounds (0.5th--99.5th percentiles of sampled frames) and
linear conversion from 14 to 8 bits. Fig.~\ref{fig:qualitative-matching} shows examples.

\subsection{Hardware Platforms}
\label{sec:hardware_platforms}
We compare two workstations (RTX 5070 Ti and RTX 5060 Ti) with two onboard platforms
(Jetson AGX Thor and Orin Nano), spanning different compute, memory, and power budgets.
Table~\ref{tab:hardware-platforms} lists specifications and power settings.\footnote{We use architecture-specific NVIDIA PyTorch containers. The RTX CPU governor is set to performance; Jetson clocks are not additionally locked. Detailed software versions and execution settings are provided on the project page.}
\begin{table}[!htbp]
\centering
\caption{Hardware platforms.}
\label{tab:hardware-platforms}
\begingroup
\fontsize{9}{10}\selectfont
\setlength{\tabcolsep}{2pt}
\renewcommand{\arraystretch}{1.0}
\resizebox{0.85\columnwidth}{!}{%
\begin{tabular*}{\columnwidth}{@{\extracolsep{\fill}}llclc@{}}
\toprule
{\renewcommand{\arraystretch}{0.9}\begin{tabular}[c]{@{}l@{}}Platform\end{tabular}} & {\renewcommand{\arraystretch}{0.9}\begin{tabular}[c]{@{}l@{}}GPU\\CUDA cores\end{tabular}} & {\renewcommand{\arraystretch}{0.9}\begin{tabular}[c]{@{}c@{}}VRAM\end{tabular}} & {\renewcommand{\arraystretch}{0.9}\begin{tabular}[c]{@{}l@{}}CPU\end{tabular}} & {\renewcommand{\arraystretch}{0.9}\begin{tabular}[c]{@{}c@{}}Power\\setting\end{tabular}} \\
\midrule
{\renewcommand{\arraystretch}{0.9}\begin{tabular}[c]{@{}l@{}}\bfseries RTX\\\bfseries 5070 Ti\end{tabular}} & {\renewcommand{\arraystretch}{0.9}\begin{tabular}[c]{@{}l@{}}Blackwell\\8,960\end{tabular}} & {\renewcommand{\arraystretch}{0.9}\begin{tabular}[c]{@{}c@{}}16 GB\\GDDR7\end{tabular}} & {\renewcommand{\arraystretch}{0.9}\begin{tabular}[c]{@{}l@{}}AMD Ryzen 9\\9900X\end{tabular}} & {\renewcommand{\arraystretch}{0.9}\begin{tabular}[c]{@{}c@{}}300 W\end{tabular}} \\
\midrule[0.25pt]
{\renewcommand{\arraystretch}{0.9}\begin{tabular}[c]{@{}l@{}}\bfseries RTX\\\bfseries 5060 Ti\end{tabular}} & {\renewcommand{\arraystretch}{0.9}\begin{tabular}[c]{@{}l@{}}Blackwell\\4,608\end{tabular}} & {\renewcommand{\arraystretch}{0.9}\begin{tabular}[c]{@{}c@{}}16 GB\\GDDR7\end{tabular}} & {\renewcommand{\arraystretch}{0.9}\begin{tabular}[c]{@{}l@{}}Intel Core Ultra 7\\265F\end{tabular}} & {\renewcommand{\arraystretch}{0.9}\begin{tabular}[c]{@{}c@{}}180 W\end{tabular}} \\
\midrule[0.25pt]
{\renewcommand{\arraystretch}{0.9}\begin{tabular}[c]{@{}l@{}}\bfseries Jetson\\\bfseries AGX Thor\end{tabular}} & {\renewcommand{\arraystretch}{0.9}\begin{tabular}[c]{@{}l@{}}Blackwell\\2,560\end{tabular}} & {\renewcommand{\arraystretch}{0.9}\begin{tabular}[c]{@{}c@{}}128 GB\\LPDDR5X\end{tabular}} & {\renewcommand{\arraystretch}{0.9}\begin{tabular}[c]{@{}l@{}}Neoverse-V3AE\\14 cores\end{tabular}} & {\renewcommand{\arraystretch}{0.9}\begin{tabular}[c]{@{}c@{}}MAXN\end{tabular}} \\
\midrule[0.25pt]
{\renewcommand{\arraystretch}{0.9}\begin{tabular}[c]{@{}l@{}}\bfseries Jetson\\\bfseries Orin Nano\end{tabular}} & {\renewcommand{\arraystretch}{0.9}\begin{tabular}[c]{@{}l@{}}Ampere\\1,024\end{tabular}} & {\renewcommand{\arraystretch}{0.9}\begin{tabular}[c]{@{}c@{}}8 GB\\LPDDR5\end{tabular}} & {\renewcommand{\arraystretch}{0.9}\begin{tabular}[c]{@{}l@{}}Cortex-A78AE\\6 cores\end{tabular}} & {\renewcommand{\arraystretch}{0.9}\begin{tabular}[c]{@{}c@{}}MAXN\\SUPER\end{tabular}} \\
\addlinespace[2pt]
\bottomrule
\end{tabular*}%
}
\endgroup
\par\vspace{4pt}
\begin{minipage}{\columnwidth}
\footnotesize\raggedright
VRAM: dedicated GPU memory on RTX; CPU--GPU shared memory on Jetson.\\
Power: RTX GPU limits; Jetson power modes.
\end{minipage}
\end{table}

\subsection{Matcher Configurations}
\label{sec:matcher_configurations}
We use public implementations and pretrained weights without additional training. MatchAnything
uses the ELoFTR-based variant (Table~\ref{tab:precision-support}).
\begin{table}[!htbp]
\centering\caption{Matchers and precision support.}
\label{tab:precision-support}
\begingroup
\definecolor{precisionSupported}{HTML}{2CA02C}
\definecolor{precisionUnsupported}{HTML}{D62728}
\fontsize{7}{8}\selectfont
\setlength{\tabcolsep}{0.7pt}
\renewcommand{\arraystretch}{1.15}
\begin{tabular*}{\columnwidth}{@{\extracolsep{\fill}}llc@{\hspace{3pt}}lccccc@{}}
\toprule
Family & Matcher & Arch. & Pub. & Backend & Native & FP32 & MP & FP16\\
\midrule
\multirow{2}{*}[-0.4pt]{{\renewcommand{\arraystretch}{0.9}\begin{tabular}[c]{@{}l@{}}Classical\\sparse\end{tabular}}} & ORB (CUDA)~\cite{rublee2011orb} & --- & ICCV'11 & OpenCV & \textcolor{precisionSupported}{$\checkmark$} & --- & --- & --- \\
 & SIFT (GPU)~\cite{lowe2004sift} & --- & IJCV'04 & COLMAP & \textcolor{precisionSupported}{$\checkmark$} & --- & --- & --- \\
\midrule
\multirow{2}{*}[-0.4pt]{{\renewcommand{\arraystretch}{0.9}\begin{tabular}[c]{@{}l@{}}Learned\\sparse\end{tabular}}} & SP+SG~\cite{detone2018superpoint,sarlin2020superglue} & CNN/GNN & CVPR'20 & PyTorch & --- & \textcolor{precisionSupported}{$\checkmark$} & \textcolor{precisionSupported}{$\checkmark$} & \textcolor{precisionSupported}{$\checkmark$} \\
 & SP+LG~\cite{detone2018superpoint,lindenberger2023lightglue} & CNN/Tr. & ICCV'23 & PyTorch & --- & \textcolor{precisionSupported}{$\checkmark$} & \textcolor{precisionSupported}{$\checkmark$} & \textcolor{precisionSupported}{$\checkmark$} \\
\midrule
\multirow{3}{*}[-1.655pt]{{\renewcommand{\arraystretch}{0.9}\begin{tabular}[c]{@{}l@{}}Learned\\semi-\\dense\end{tabular}}} & LoFTR~\cite{sun2021loftr} & CNN/Tr. & CVPR'21 & PyTorch & --- & \textcolor{precisionSupported}{$\checkmark$} & \textcolor{precisionSupported}{$\checkmark$} & \textcolor{precisionSupported}{$\checkmark$} \\
 & ELoFTR~\cite{wang2024eloftr} & CNN/Tr. & CVPR'24 & PyTorch & --- & \textcolor{precisionSupported}{$\checkmark$} & \textcolor{precisionSupported}{$\checkmark$} & \textcolor{precisionSupported}{$\checkmark$} \\
 & {\renewcommand{\arraystretch}{0.9}\begin{tabular}[c]{@{}l@{}}MatchAnything~\cite{He2025MatchAnything}\\(ELoFTR)\end{tabular}} & CNN/Tr. & TPAMI'26 & PyTorch & --- & \textcolor{precisionSupported}{$\checkmark$} & \textcolor{precisionSupported}{$\checkmark$} & \textcolor{precisionSupported}{$\checkmark$} \\
\midrule
\multirow{2}{*}[-0.4pt]{{\renewcommand{\arraystretch}{0.9}\begin{tabular}[c]{@{}l@{}}Learned\\dense\end{tabular}}} & DKM~\cite{edstedt2023dkm} & CNN & CVPR'23 & PyTorch & --- & \textcolor{precisionSupported}{$\checkmark$} & \textcolor{precisionSupported}{$\checkmark$} & \textcolor{precisionUnsupported}{$\times$} \\
 & MASt3R~\cite{leroy2024mast3r} & ViT & ECCV'24 & PyTorch & --- & \textcolor{precisionSupported}{$\checkmark$} & \textcolor{precisionSupported}{$\checkmark$} & \textcolor{precisionUnsupported}{$\times$} \\
\bottomrule
\end{tabular*}
\endgroup
\par\vspace{4pt}
\begin{minipage}{\columnwidth}
\footnotesize\raggedright
SP: SuperPoint; SG: SuperGlue; LG: LightGlue;\\
Tr.: Transformer; MP: mixed precision.\\
\textcolor[HTML]{2CA02C}{$\checkmark$}: evaluated; \textcolor[HTML]{D62728}{$\times$}: unsupported; ---: not applicable.
\end{minipage}
\end{table}

 Matching settings are fixed across
datasets and platforms. Inputs are resized with bilinear interpolation while preserving aspect ratio;
correspondences are restored to original coordinates for a common pose estimator.
FP32 and FP16 use the corresponding weight and input types without autocasting; MP retains FP32
weights with FP16 autocasting. TF32 is disabled for all learned pipelines. The evaluation grid and
supported modes are given in Section~\ref{sec:matchercompass} and Table~\ref{tab:precision-support}.

\subsection{Matching Accuracy via Pose Estimation}
\label{sec:pose_evaluation}
We evaluate all matchers on RTX 5070 Ti following the pose evaluation protocol of
SuperGlue~\cite{sarlin2020superglue}.

\subsection{Runtime, Memory, and Energy Measurement}
\label{sec:computational_measurements}
\begin{figure}[!t]
\centering
\resizebox{0.9\columnwidth}{!}{%
\includegraphics{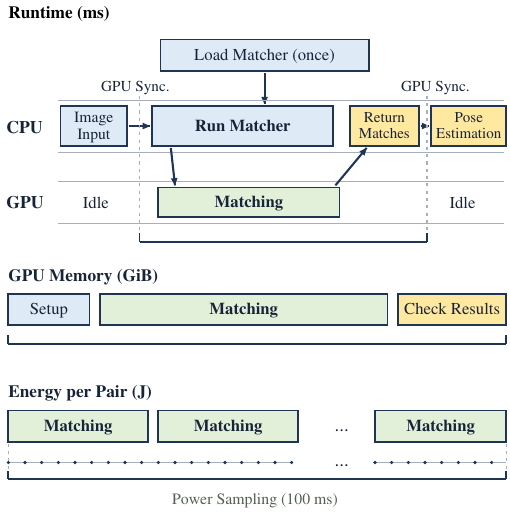}%
}
\caption{Runtime, GPU memory, and energy measurement. GPU synchronization bounds the timed matcher call, excluding model loading and pose estimation. Runtime is measured separately from memory and energy to avoid tracing overhead.}
\label{fig:measurement-timeline}
\end{figure}

For deployment-aware selection, we measure runtime, peak GPU memory, and energy per pair
to address engineers' latency, memory, and battery constraints in field robotics.
All configurations use the same 16 pairs (four per task) across platforms.
Fig.~\ref{fig:measurement-timeline} summarizes the three measurement scopes. GPU synchronization
bounds the timed matcher call, excluding model loading and pose estimation. After warm-up,
runtime is the median of three calls per pair; task-level results are medians over four pairs.
CUPTI tracks peak CUDA allocation, including model memory and buffers.\footnote{Each run begins with 3~s of idle time and 42 warm-up calls. Energy measurements repeat each pair for at least 6~s and five calls, sampling power at a target interval of 100~ms.}

\begin{figure*}[!t]
\centering
\includegraphics[width=\textwidth]{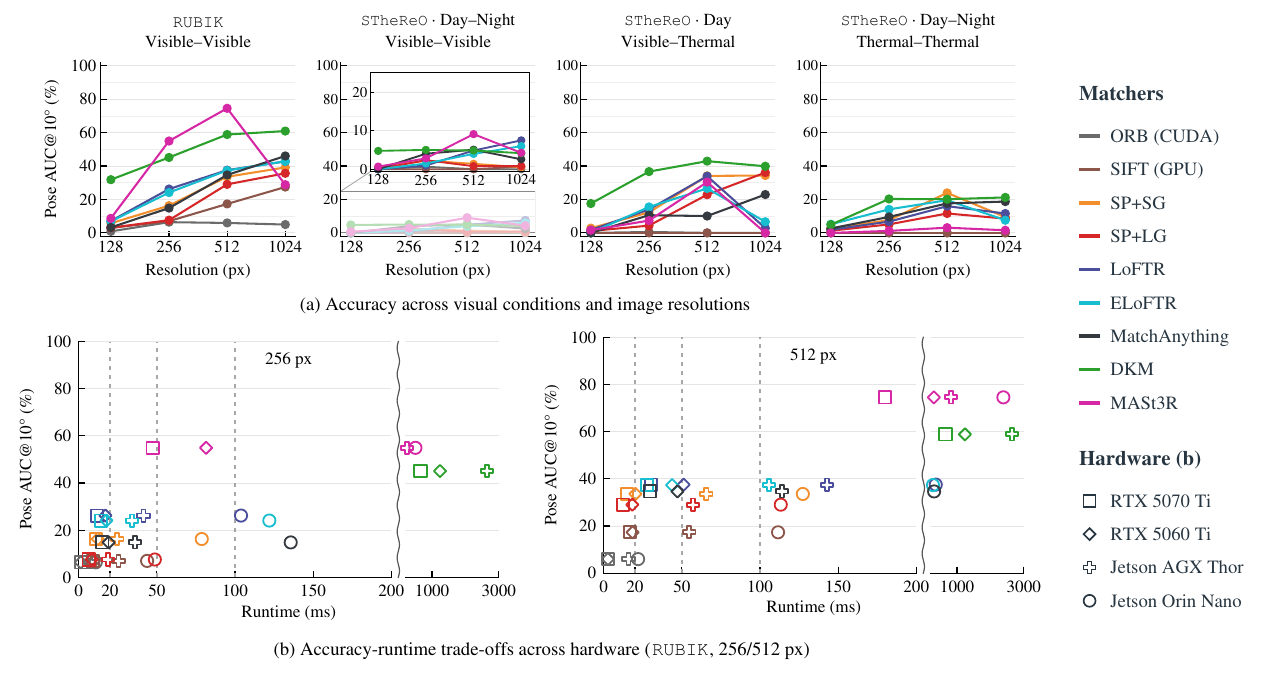}
\caption{Matching accuracy and deployment trade-offs. (a) Pose AUC@10$^\circ$ over 100 pairs per task and resolution on RTX 5070 Ti. (b) Accuracy versus median runtime over four pairs per platform on \texttt{RUBIK} at 256~px (left) and 512~px (right). Dashed lines indicate 20/50/100~ms; axes break at 200~ms. DKM on Orin Nano is omitted due to OOM. Learned matchers use FP32; ORB (CUDA)/SIFT (GPU) use Native. Matcher colors are shared; hardware symbols apply to (b). SP: SuperPoint; SG: SuperGlue; LG: LightGlue.}
\label{fig:accuracy-overview}
\end{figure*}

Energy per pair is integrated power divided by completed calls,
$E_{\mathrm{pair}}=N^{-1}\sum_i P_i\Delta t_i$, including idle power.
Power is measured using \texttt{nvidia-smi} on RTX and \texttt{tegrastats} on Jetson.
The measurement scope differs: RTX reports GPU board power, Thor the GPU rail, and Orin Nano
the combined CPU, GPU, and CV rail. These scopes must be considered when comparing energy across platforms.

\begin{table*}[!t]
\centering
\captionsetup{justification=centering}
\caption{Hardware-dependent runtime, peak GPU memory, and energy per pair on \texttt{RUBIK} at 512~px,\\
using FP32 for learned matchers and Native for ORB (CUDA)/SIFT (GPU).}
\label{tab:hardware-cost-comparison}
\begingroup
\definecolor{matcher0}{HTML}{6B6B6B}
\definecolor{matcher1}{HTML}{8C564B}
\definecolor{matcher2}{HTML}{F28E2B}
\definecolor{matcher3}{HTML}{D62728}
\definecolor{matcher4}{HTML}{4B4E9A}
\definecolor{matcher5}{HTML}{17BECF}
\definecolor{matcher6}{HTML}{343A40}
\definecolor{matcher7}{HTML}{2CA02C}
\definecolor{matcher8}{HTML}{D627A5}
\fontsize{9}{10}\selectfont
\setlength{\tabcolsep}{2pt}\setlength{\arrayrulewidth}{0.25pt}
\renewcommand{\arraystretch}{0.85}
\resizebox{0.9\textwidth}{!}{%
\begin{tabularx}{\textwidth}{@{}l *{3}{>{\raggedleft\arraybackslash}X} | *{3}{>{\raggedleft\arraybackslash}X} | *{3}{>{\raggedleft\arraybackslash}X} | *{3}{>{\raggedleft\arraybackslash}X} @{}}
\toprule
 & \multicolumn{6}{c|}{Desktop GPUs} & \multicolumn{6}{c}{Embedded Platforms} \\
\cmidrule(lr){2-7}\cmidrule(lr){8-13}
\raisebox{-14.67pt}[0pt][0pt]{Matcher} & \multicolumn{3}{c|}{RTX 5070 Ti} & \multicolumn{3}{c|}{RTX 5060 Ti} & \multicolumn{3}{c|}{Jetson AGX Thor} & \multicolumn{3}{c@{}}{Jetson Orin Nano} \\
\cmidrule(lr){2-4}\cmidrule(lr){5-7}\cmidrule(lr){8-10}\cmidrule(lr){11-13}
 & \multicolumn{1}{c}{Runtime} & \multicolumn{1}{c}{GPU} & \multicolumn{1}{c|}{Energy} & \multicolumn{1}{c}{Runtime} & \multicolumn{1}{c}{GPU} & \multicolumn{1}{c|}{Energy} & \multicolumn{1}{c}{Runtime} & \multicolumn{1}{c}{GPU} & \multicolumn{1}{c|}{Energy} & \multicolumn{1}{c}{Runtime} & \multicolumn{1}{c}{GPU} & \multicolumn{1}{c@{}}{Energy} \\
 & \multicolumn{1}{c}{} & \multicolumn{1}{c}{memory} & \multicolumn{1}{c|}{per pair} & \multicolumn{1}{c}{} & \multicolumn{1}{c}{memory} & \multicolumn{1}{c|}{per pair} & \multicolumn{1}{c}{} & \multicolumn{1}{c}{memory} & \multicolumn{1}{c|}{per pair} & \multicolumn{1}{c}{} & \multicolumn{1}{c}{memory} & \multicolumn{1}{c@{}}{per pair} \\
 & \multicolumn{1}{c}{(ms)} & \multicolumn{1}{c}{(GiB)} & \multicolumn{1}{c|}{(J)} & \multicolumn{1}{c}{(ms)} & \multicolumn{1}{c}{(GiB)} & \multicolumn{1}{c|}{(J)} & \multicolumn{1}{c}{(ms)} & \multicolumn{1}{c}{(GiB)} & \multicolumn{1}{c|}{(J)} & \multicolumn{1}{c}{(ms)} & \multicolumn{1}{c}{(GiB)} & \multicolumn{1}{c@{}}{(J)} \\
\midrule
\rowcolor{matcher0!10}
\raisebox{-3.35pt}[0pt][0pt]{\textbf{ORB (CUDA)}~\cite{rublee2011orb}} & {\renewcommand{\arraystretch}{0.75}\begin{tabular}[t]{@{}r@{}}2.6\\{\fontsize{7.2}{8}\selectfont (0.17\texttimes{})}\end{tabular}} & {\renewcommand{\arraystretch}{0.75}\begin{tabular}[t]{@{}r@{}}0.003\\{\fontsize{7.2}{8}\selectfont (0.01\texttimes{})}\end{tabular}} & {\renewcommand{\arraystretch}{0.75}\begin{tabular}[t]{@{}r@{}}0.18\\{\fontsize{7.2}{8}\selectfont (0.06\texttimes{})}\end{tabular}} & {\renewcommand{\arraystretch}{0.75}\begin{tabular}[t]{@{}r@{}}2.7\\{\fontsize{7.2}{8}\selectfont (0.13\texttimes{})}\end{tabular}} & {\renewcommand{\arraystretch}{0.75}\begin{tabular}[t]{@{}r@{}}0.003\\{\fontsize{7.2}{8}\selectfont (0.01\texttimes{})}\end{tabular}} & {\renewcommand{\arraystretch}{0.75}\begin{tabular}[t]{@{}r@{}}0.09\\{\fontsize{7.2}{8}\selectfont (0.03\texttimes{})}\end{tabular}} & {\renewcommand{\arraystretch}{0.75}\begin{tabular}[t]{@{}r@{}}15.8\\{\fontsize{7.2}{8}\selectfont (0.24\texttimes{})}\end{tabular}} & {\renewcommand{\arraystretch}{0.75}\begin{tabular}[t]{@{}r@{}}0.003\\{\fontsize{7.2}{8}\selectfont (0.0045\texttimes{})}\end{tabular}} & {\renewcommand{\arraystretch}{0.75}\begin{tabular}[t]{@{}r@{}}0.05\\{\fontsize{7.2}{8}\selectfont (0.02\texttimes{})}\end{tabular}} & {\renewcommand{\arraystretch}{0.75}\begin{tabular}[t]{@{}r@{}}22.0\\{\fontsize{7.2}{8}\selectfont (0.17\texttimes{})}\end{tabular}} & {\renewcommand{\arraystretch}{0.75}\begin{tabular}[t]{@{}r@{}}0.003\\{\fontsize{7.2}{8}\selectfont (0.01\texttimes{})}\end{tabular}} & {\renewcommand{\arraystretch}{0.75}\begin{tabular}[t]{@{}r@{}}0.03\\{\fontsize{7.2}{8}\selectfont (0.03\texttimes{})}\end{tabular}} \\
\rowcolor{matcher1!10}
\raisebox{-3.35pt}[0pt][0pt]{\textbf{SIFT (GPU)}~\cite{lowe2004sift}} & {\renewcommand{\arraystretch}{0.75}\begin{tabular}[t]{@{}r@{}}16.7\\{\fontsize{7.2}{8}\selectfont (1.11\texttimes{})}\end{tabular}} & {\renewcommand{\arraystretch}{0.75}\begin{tabular}[t]{@{}r@{}}0.155\\{\fontsize{7.2}{8}\selectfont (0.47\texttimes{})}\end{tabular}} & {\renewcommand{\arraystretch}{0.75}\begin{tabular}[t]{@{}r@{}}1.35\\{\fontsize{7.2}{8}\selectfont (0.45\texttimes{})}\end{tabular}} & {\renewcommand{\arraystretch}{0.75}\begin{tabular}[t]{@{}r@{}}18.5\\{\fontsize{7.2}{8}\selectfont (0.91\texttimes{})}\end{tabular}} & {\renewcommand{\arraystretch}{0.75}\begin{tabular}[t]{@{}r@{}}0.155\\{\fontsize{7.2}{8}\selectfont (0.47\texttimes{})}\end{tabular}} & {\renewcommand{\arraystretch}{0.75}\begin{tabular}[t]{@{}r@{}}0.72\\{\fontsize{7.2}{8}\selectfont (0.28\texttimes{})}\end{tabular}} & {\renewcommand{\arraystretch}{0.75}\begin{tabular}[t]{@{}r@{}}54.8\\{\fontsize{7.2}{8}\selectfont (0.83\texttimes{})}\end{tabular}} & {\renewcommand{\arraystretch}{0.75}\begin{tabular}[t]{@{}r@{}}0.155\\{\fontsize{7.2}{8}\selectfont (0.21\texttimes{})}\end{tabular}} & {\renewcommand{\arraystretch}{0.75}\begin{tabular}[t]{@{}r@{}}0.51\\{\fontsize{7.2}{8}\selectfont (0.25\texttimes{})}\end{tabular}} & {\renewcommand{\arraystretch}{0.75}\begin{tabular}[t]{@{}r@{}}111.6\\{\fontsize{7.2}{8}\selectfont (0.88\texttimes{})}\end{tabular}} & {\renewcommand{\arraystretch}{0.75}\begin{tabular}[t]{@{}r@{}}0.155\\{\fontsize{7.2}{8}\selectfont (0.64\texttimes{})}\end{tabular}} & {\renewcommand{\arraystretch}{0.75}\begin{tabular}[t]{@{}r@{}}0.30\\{\fontsize{7.2}{8}\selectfont (0.30\texttimes{})}\end{tabular}} \\
\rowcolor{matcher2!10}
\raisebox{-3.35pt}[0pt][0pt]{\textbf{SP+SG}~\cite{detone2018superpoint,sarlin2020superglue}} & {\renewcommand{\arraystretch}{0.75}\begin{tabular}[t]{@{}r@{}}15.0\\{\fontsize{7.2}{8}\selectfont (1.00\texttimes{})}\end{tabular}} & {\renewcommand{\arraystretch}{0.75}\begin{tabular}[t]{@{}r@{}}0.330\\{\fontsize{7.2}{8}\selectfont (1.00\texttimes{})}\end{tabular}} & {\renewcommand{\arraystretch}{0.75}\begin{tabular}[t]{@{}r@{}}3.00\\{\fontsize{7.2}{8}\selectfont (1.00\texttimes{})}\end{tabular}} & {\renewcommand{\arraystretch}{0.75}\begin{tabular}[t]{@{}r@{}}20.3\\{\fontsize{7.2}{8}\selectfont (1.00\texttimes{})}\end{tabular}} & {\renewcommand{\arraystretch}{0.75}\begin{tabular}[t]{@{}r@{}}0.330\\{\fontsize{7.2}{8}\selectfont (1.00\texttimes{})}\end{tabular}} & {\renewcommand{\arraystretch}{0.75}\begin{tabular}[t]{@{}r@{}}2.56\\{\fontsize{7.2}{8}\selectfont (1.00\texttimes{})}\end{tabular}} & {\renewcommand{\arraystretch}{0.75}\begin{tabular}[t]{@{}r@{}}65.7\\{\fontsize{7.2}{8}\selectfont (1.00\texttimes{})}\end{tabular}} & {\renewcommand{\arraystretch}{0.75}\begin{tabular}[t]{@{}r@{}}0.731\\{\fontsize{7.2}{8}\selectfont (1.00\texttimes{})}\end{tabular}} & {\renewcommand{\arraystretch}{0.75}\begin{tabular}[t]{@{}r@{}}2.04\\{\fontsize{7.2}{8}\selectfont (1.00\texttimes{})}\end{tabular}} & {\renewcommand{\arraystretch}{0.75}\begin{tabular}[t]{@{}r@{}}127.4\\{\fontsize{7.2}{8}\selectfont (1.00\texttimes{})}\end{tabular}} & {\renewcommand{\arraystretch}{0.75}\begin{tabular}[t]{@{}r@{}}0.244\\{\fontsize{7.2}{8}\selectfont (1.00\texttimes{})}\end{tabular}} & {\renewcommand{\arraystretch}{0.75}\begin{tabular}[t]{@{}r@{}}1.02\\{\fontsize{7.2}{8}\selectfont (1.00\texttimes{})}\end{tabular}} \\
\rowcolor{matcher3!10}
\raisebox{-3.35pt}[0pt][0pt]{\textbf{SP+LG}~\cite{detone2018superpoint,lindenberger2023lightglue}} & {\renewcommand{\arraystretch}{0.75}\begin{tabular}[t]{@{}r@{}}12.2\\{\fontsize{7.2}{8}\selectfont (0.81\texttimes{})}\end{tabular}} & {\renewcommand{\arraystretch}{0.75}\begin{tabular}[t]{@{}r@{}}0.332\\{\fontsize{7.2}{8}\selectfont (1.01\texttimes{})}\end{tabular}} & {\renewcommand{\arraystretch}{0.75}\begin{tabular}[t]{@{}r@{}}2.73\\{\fontsize{7.2}{8}\selectfont (0.91\texttimes{})}\end{tabular}} & {\renewcommand{\arraystretch}{0.75}\begin{tabular}[t]{@{}r@{}}18.3\\{\fontsize{7.2}{8}\selectfont (0.90\texttimes{})}\end{tabular}} & {\renewcommand{\arraystretch}{0.75}\begin{tabular}[t]{@{}r@{}}0.332\\{\fontsize{7.2}{8}\selectfont (1.01\texttimes{})}\end{tabular}} & {\renewcommand{\arraystretch}{0.75}\begin{tabular}[t]{@{}r@{}}2.58\\{\fontsize{7.2}{8}\selectfont (1.01\texttimes{})}\end{tabular}} & {\renewcommand{\arraystretch}{0.75}\begin{tabular}[t]{@{}r@{}}57.2\\{\fontsize{7.2}{8}\selectfont (0.87\texttimes{})}\end{tabular}} & {\renewcommand{\arraystretch}{0.75}\begin{tabular}[t]{@{}r@{}}0.733\\{\fontsize{7.2}{8}\selectfont (1.00\texttimes{})}\end{tabular}} & {\renewcommand{\arraystretch}{0.75}\begin{tabular}[t]{@{}r@{}}1.91\\{\fontsize{7.2}{8}\selectfont (0.94\texttimes{})}\end{tabular}} & {\renewcommand{\arraystretch}{0.75}\begin{tabular}[t]{@{}r@{}}113.3\\{\fontsize{7.2}{8}\selectfont (0.89\texttimes{})}\end{tabular}} & {\renewcommand{\arraystretch}{0.75}\begin{tabular}[t]{@{}r@{}}0.246\\{\fontsize{7.2}{8}\selectfont (1.01\texttimes{})}\end{tabular}} & {\renewcommand{\arraystretch}{0.75}\begin{tabular}[t]{@{}r@{}}0.96\\{\fontsize{7.2}{8}\selectfont (0.94\texttimes{})}\end{tabular}} \\
\rowcolor{matcher4!10}
\raisebox{-3.35pt}[0pt][0pt]{\textbf{LoFTR}~\cite{sun2021loftr}} & {\renewcommand{\arraystretch}{0.75}\begin{tabular}[t]{@{}r@{}}30.1\\{\fontsize{7.2}{8}\selectfont (2.00\texttimes{})}\end{tabular}} & {\renewcommand{\arraystretch}{0.75}\begin{tabular}[t]{@{}r@{}}1.471\\{\fontsize{7.2}{8}\selectfont (4.45\texttimes{})}\end{tabular}} & {\renewcommand{\arraystretch}{0.75}\begin{tabular}[t]{@{}r@{}}8.39\\{\fontsize{7.2}{8}\selectfont (2.79\texttimes{})}\end{tabular}} & {\renewcommand{\arraystretch}{0.75}\begin{tabular}[t]{@{}r@{}}51.2\\{\fontsize{7.2}{8}\selectfont (2.52\texttimes{})}\end{tabular}} & {\renewcommand{\arraystretch}{0.75}\begin{tabular}[t]{@{}r@{}}1.354\\{\fontsize{7.2}{8}\selectfont (4.10\texttimes{})}\end{tabular}} & {\renewcommand{\arraystretch}{0.75}\begin{tabular}[t]{@{}r@{}}8.14\\{\fontsize{7.2}{8}\selectfont (3.17\texttimes{})}\end{tabular}} & {\renewcommand{\arraystretch}{0.75}\begin{tabular}[t]{@{}r@{}}142.9\\{\fontsize{7.2}{8}\selectfont (2.18\texttimes{})}\end{tabular}} & {\renewcommand{\arraystretch}{0.75}\begin{tabular}[t]{@{}r@{}}1.360\\{\fontsize{7.2}{8}\selectfont (1.86\texttimes{})}\end{tabular}} & {\renewcommand{\arraystretch}{0.75}\begin{tabular}[t]{@{}r@{}}5.67\\{\fontsize{7.2}{8}\selectfont (2.79\texttimes{})}\end{tabular}} & {\renewcommand{\arraystretch}{0.75}\begin{tabular}[t]{@{}r@{}}360.1\\{\fontsize{7.2}{8}\selectfont (2.83\texttimes{})}\end{tabular}} & {\renewcommand{\arraystretch}{0.75}\begin{tabular}[t]{@{}r@{}}1.352\\{\fontsize{7.2}{8}\selectfont (5.53\texttimes{})}\end{tabular}} & {\renewcommand{\arraystretch}{0.75}\begin{tabular}[t]{@{}r@{}}3.53\\{\fontsize{7.2}{8}\selectfont (3.45\texttimes{})}\end{tabular}} \\
\rowcolor{matcher5!10}
\raisebox{-3.35pt}[0pt][0pt]{\textbf{ELoFTR}~\cite{wang2024eloftr}} & {\renewcommand{\arraystretch}{0.75}\begin{tabular}[t]{@{}r@{}}27.8\\{\fontsize{7.2}{8}\selectfont (1.85\texttimes{})}\end{tabular}} & {\renewcommand{\arraystretch}{0.75}\begin{tabular}[t]{@{}r@{}}2.190\\{\fontsize{7.2}{8}\selectfont (6.63\texttimes{})}\end{tabular}} & {\renewcommand{\arraystretch}{0.75}\begin{tabular}[t]{@{}r@{}}6.42\\{\fontsize{7.2}{8}\selectfont (2.14\texttimes{})}\end{tabular}} & {\renewcommand{\arraystretch}{0.75}\begin{tabular}[t]{@{}r@{}}43.8\\{\fontsize{7.2}{8}\selectfont (2.16\texttimes{})}\end{tabular}} & {\renewcommand{\arraystretch}{0.75}\begin{tabular}[t]{@{}r@{}}2.049\\{\fontsize{7.2}{8}\selectfont (6.21\texttimes{})}\end{tabular}} & {\renewcommand{\arraystretch}{0.75}\begin{tabular}[t]{@{}r@{}}5.99\\{\fontsize{7.2}{8}\selectfont (2.34\texttimes{})}\end{tabular}} & {\renewcommand{\arraystretch}{0.75}\begin{tabular}[t]{@{}r@{}}105.6\\{\fontsize{7.2}{8}\selectfont (1.61\texttimes{})}\end{tabular}} & {\renewcommand{\arraystretch}{0.75}\begin{tabular}[t]{@{}r@{}}2.112\\{\fontsize{7.2}{8}\selectfont (2.89\texttimes{})}\end{tabular}} & {\renewcommand{\arraystretch}{0.75}\begin{tabular}[t]{@{}r@{}}3.87\\{\fontsize{7.2}{8}\selectfont (1.90\texttimes{})}\end{tabular}} & {\renewcommand{\arraystretch}{0.75}\begin{tabular}[t]{@{}r@{}}277.2\\{\fontsize{7.2}{8}\selectfont (2.18\texttimes{})}\end{tabular}} & {\renewcommand{\arraystretch}{0.75}\begin{tabular}[t]{@{}r@{}}2.111\\{\fontsize{7.2}{8}\selectfont (8.64\texttimes{})}\end{tabular}} & {\renewcommand{\arraystretch}{0.75}\begin{tabular}[t]{@{}r@{}}2.49\\{\fontsize{7.2}{8}\selectfont (2.44\texttimes{})}\end{tabular}} \\
\rowcolor{matcher6!12}
\raisebox{-3.35pt}[0pt][0pt]{\textbf{MatchAnything}~\cite{He2025MatchAnything}} & {\renewcommand{\arraystretch}{0.75}\begin{tabular}[t]{@{}r@{}}29.6\\{\fontsize{7.2}{8}\selectfont (1.97\texttimes{})}\end{tabular}} & {\renewcommand{\arraystretch}{0.75}\begin{tabular}[t]{@{}r@{}}2.130\\{\fontsize{7.2}{8}\selectfont (6.45\texttimes{})}\end{tabular}} & {\renewcommand{\arraystretch}{0.75}\begin{tabular}[t]{@{}r@{}}7.01\\{\fontsize{7.2}{8}\selectfont (2.33\texttimes{})}\end{tabular}} & {\renewcommand{\arraystretch}{0.75}\begin{tabular}[t]{@{}r@{}}47.2\\{\fontsize{7.2}{8}\selectfont (2.33\texttimes{})}\end{tabular}} & {\renewcommand{\arraystretch}{0.75}\begin{tabular}[t]{@{}r@{}}2.130\\{\fontsize{7.2}{8}\selectfont (6.45\texttimes{})}\end{tabular}} & {\renewcommand{\arraystretch}{0.75}\begin{tabular}[t]{@{}r@{}}6.36\\{\fontsize{7.2}{8}\selectfont (2.48\texttimes{})}\end{tabular}} & {\renewcommand{\arraystretch}{0.75}\begin{tabular}[t]{@{}r@{}}114.0\\{\fontsize{7.2}{8}\selectfont (1.73\texttimes{})}\end{tabular}} & {\renewcommand{\arraystretch}{0.75}\begin{tabular}[t]{@{}r@{}}2.130\\{\fontsize{7.2}{8}\selectfont (2.91\texttimes{})}\end{tabular}} & {\renewcommand{\arraystretch}{0.75}\begin{tabular}[t]{@{}r@{}}4.07\\{\fontsize{7.2}{8}\selectfont (2.00\texttimes{})}\end{tabular}} & {\renewcommand{\arraystretch}{0.75}\begin{tabular}[t]{@{}r@{}}310.3\\{\fontsize{7.2}{8}\selectfont (2.44\texttimes{})}\end{tabular}} & {\renewcommand{\arraystretch}{0.75}\begin{tabular}[t]{@{}r@{}}2.129\\{\fontsize{7.2}{8}\selectfont (8.71\texttimes{})}\end{tabular}} & {\renewcommand{\arraystretch}{0.75}\begin{tabular}[t]{@{}r@{}}2.74\\{\fontsize{7.2}{8}\selectfont (2.68\texttimes{})}\end{tabular}} \\
\rowcolor{matcher7!10}
\raisebox{-3.35pt}[0pt][0pt]{\textbf{DKM}~\cite{edstedt2023dkm}} & {\renewcommand{\arraystretch}{0.75}\begin{tabular}[t]{@{}r@{}}651.9\\{\fontsize{7.2}{8}\selectfont (43.36\texttimes{})}\end{tabular}} & {\renewcommand{\arraystretch}{0.75}\begin{tabular}[t]{@{}r@{}}13.442\\{\fontsize{7.2}{8}\selectfont (40.70\texttimes{})}\end{tabular}} & {\renewcommand{\arraystretch}{0.75}\begin{tabular}[t]{@{}r@{}}163.46\\{\fontsize{7.2}{8}\selectfont (54.43\texttimes{})}\end{tabular}} & {\renewcommand{\arraystretch}{0.75}\begin{tabular}[t]{@{}r@{}}1236.8\\{\fontsize{7.2}{8}\selectfont (61.03\texttimes{})}\end{tabular}} & {\renewcommand{\arraystretch}{0.75}\begin{tabular}[t]{@{}r@{}}13.975\\{\fontsize{7.2}{8}\selectfont (42.31\texttimes{})}\end{tabular}} & {\renewcommand{\arraystretch}{0.75}\begin{tabular}[t]{@{}r@{}}158.82\\{\fontsize{7.2}{8}\selectfont (61.92\texttimes{})}\end{tabular}} & {\renewcommand{\arraystretch}{0.75}\begin{tabular}[t]{@{}r@{}}2649.4\\{\fontsize{7.2}{8}\selectfont (40.33\texttimes{})}\end{tabular}} & {\renewcommand{\arraystretch}{0.75}\begin{tabular}[t]{@{}r@{}}13.975\\{\fontsize{7.2}{8}\selectfont (19.13\texttimes{})}\end{tabular}} & {\renewcommand{\arraystretch}{0.75}\begin{tabular}[t]{@{}r@{}}94.81\\{\fontsize{7.2}{8}\selectfont (46.54\texttimes{})}\end{tabular}} & {\renewcommand{\arraystretch}{0.75}\begin{tabular}[t]{@{}r@{}}OOM\\{\fontsize{7.2}{8}\selectfont \phantom{(1.00\texttimes{})}}\end{tabular}} & {\renewcommand{\arraystretch}{0.75}\begin{tabular}[t]{@{}r@{}}OOM\\{\fontsize{7.2}{8}\selectfont \phantom{(1.00\texttimes{})}}\end{tabular}} & {\renewcommand{\arraystretch}{0.75}\begin{tabular}[t]{@{}r@{}}OOM\\{\fontsize{7.2}{8}\selectfont \phantom{(1.00\texttimes{})}}\end{tabular}} \\
\rowcolor{matcher8!10}
\raisebox{-3.35pt}[0pt][0pt]{\textbf{MASt3R}~\cite{leroy2024mast3r}} & {\renewcommand{\arraystretch}{0.75}\begin{tabular}[t]{@{}r@{}}179.9\\{\fontsize{7.2}{8}\selectfont (11.97\texttimes{})}\end{tabular}} & {\renewcommand{\arraystretch}{0.75}\begin{tabular}[t]{@{}r@{}}4.709\\{\fontsize{7.2}{8}\selectfont (14.26\texttimes{})}\end{tabular}} & {\renewcommand{\arraystretch}{0.75}\begin{tabular}[t]{@{}r@{}}44.21\\{\fontsize{7.2}{8}\selectfont (14.72\texttimes{})}\end{tabular}} & {\renewcommand{\arraystretch}{0.75}\begin{tabular}[t]{@{}r@{}}304.6\\{\fontsize{7.2}{8}\selectfont (15.03\texttimes{})}\end{tabular}} & {\renewcommand{\arraystretch}{0.75}\begin{tabular}[t]{@{}r@{}}4.926\\{\fontsize{7.2}{8}\selectfont (14.91\texttimes{})}\end{tabular}} & {\renewcommand{\arraystretch}{0.75}\begin{tabular}[t]{@{}r@{}}43.15\\{\fontsize{7.2}{8}\selectfont (16.83\texttimes{})}\end{tabular}} & {\renewcommand{\arraystretch}{0.75}\begin{tabular}[t]{@{}r@{}}816.9\\{\fontsize{7.2}{8}\selectfont (12.44\texttimes{})}\end{tabular}} & {\renewcommand{\arraystretch}{0.75}\begin{tabular}[t]{@{}r@{}}5.952\\{\fontsize{7.2}{8}\selectfont (8.15\texttimes{})}\end{tabular}} & {\renewcommand{\arraystretch}{0.75}\begin{tabular}[t]{@{}r@{}}27.59\\{\fontsize{7.2}{8}\selectfont (13.54\texttimes{})}\end{tabular}} & {\renewcommand{\arraystretch}{0.75}\begin{tabular}[t]{@{}r@{}}2388.7\\{\fontsize{7.2}{8}\selectfont (18.75\texttimes{})}\end{tabular}} & {\renewcommand{\arraystretch}{0.75}\begin{tabular}[t]{@{}r@{}}4.100\\{\fontsize{7.2}{8}\selectfont (16.78\texttimes{})}\end{tabular}} & {\renewcommand{\arraystretch}{0.75}\begin{tabular}[t]{@{}r@{}}22.99\\{\fontsize{7.2}{8}\selectfont (22.47\texttimes{})}\end{tabular}} \\
\bottomrule
\end{tabularx}%
}
\endgroup
\par\vspace{4pt}
\begin{minipage}{0.9\textwidth}
\footnotesize\raggedright
Parentheses: ratios to SP+SG on each platform. OOM: out of memory.\\
SP+SG: SuperPoint+SuperGlue. Row colors follow Fig.~\ref{fig:accuracy-overview}.
\end{minipage}

\par\vspace{12pt}
\centering
\captionsetup{justification=centering}
\caption{Effect of numerical precision (FP32, MP, and FP16) on pose AUC and runtime\\ on \texttt{RUBIK} at 512~px. ORB (CUDA)/SIFT (GPU) use Native.}
\label{tab:precision-accuracy-runtime}
\begingroup
\definecolor{matcher0}{HTML}{6B6B6B}
\definecolor{matcher1}{HTML}{8C564B}
\definecolor{matcher2}{HTML}{F28E2B}
\definecolor{matcher3}{HTML}{D62728}
\definecolor{matcher4}{HTML}{4B4E9A}
\definecolor{matcher5}{HTML}{17BECF}
\definecolor{matcher6}{HTML}{343A40}
\definecolor{matcher7}{HTML}{2CA02C}
\definecolor{matcher8}{HTML}{D627A5}
\fontsize{9}{10}\selectfont
\setlength{\tabcolsep}{3.5pt}\setlength{\arrayrulewidth}{0.25pt}
\renewcommand{\arraystretch}{1.0}
\newsavebox{\precisiontablebox}\sbox{\precisiontablebox}{\scalebox{0.9}{%
\begin{tabularx}{\textwidth}{@{} l@{\hspace{6pt}} l *{3}{>{\raggedleft\arraybackslash}p{22pt}} | *{4}{>{\raggedleft\arraybackslash}X} @{}}
\toprule
\multirow{3}{*}[-5pt]{Matcher} & \multirow{3}{*}[-5pt]{Precision} & \multicolumn{3}{c|}{\multirow{2}{*}{Pose AUC (\%)}} & \multicolumn{4}{c}{Runtime (ms)} \\
\cmidrule(lr){6-9}
 & & \multicolumn{3}{c|}{} & \multicolumn{2}{c}{Desktop GPUs} & \multicolumn{2}{c}{Embedded Platforms} \\
\cmidrule(lr){3-5}\cmidrule(lr){6-7}\cmidrule(lr){8-9}
 & & 5\textdegree{} & 10\textdegree{} & 20\textdegree{} & \multicolumn{1}{c}{\shortstack{RTX\\5070 Ti}} & \multicolumn{1}{c}{\shortstack{RTX\\5060 Ti}} & \multicolumn{1}{c}{\shortstack{Jetson\\AGX Thor}} & \multicolumn{1}{c}{\shortstack{Jetson\\Orin Nano}} \\
\midrule
\rowcolor{matcher0!10}
\textbf{ORB (CUDA)}~\cite{rublee2011orb} & Native & 2.6 & 6.0 & 10.8 & 2.6\hspace{1pt}{\fontsize{7.2}{8}\selectfont\settowidth{\dimen0}{(\textminus{}31\%)}\makebox[\dimen0][r]{}} & 2.7\hspace{1pt}{\fontsize{7.2}{8}\selectfont\settowidth{\dimen0}{(\textminus{}31\%)}\makebox[\dimen0][r]{}} & 15.8\hspace{1pt}{\fontsize{7.2}{8}\selectfont\settowidth{\dimen0}{(\textminus{}43\%)}\makebox[\dimen0][r]{}} & 22.0\hspace{1pt}{\fontsize{7.2}{8}\selectfont\settowidth{\dimen0}{(\textminus{}28\%)}\makebox[\dimen0][r]{}} \\
\rowcolor{matcher1!10}
\textbf{SIFT (GPU)}~\cite{lowe2004sift} & Native & 11.3 & 17.3 & 26.0 & 16.7\hspace{1pt}{\fontsize{7.2}{8}\selectfont\settowidth{\dimen0}{(\textminus{}31\%)}\makebox[\dimen0][r]{}} & 18.5\hspace{1pt}{\fontsize{7.2}{8}\selectfont\settowidth{\dimen0}{(\textminus{}31\%)}\makebox[\dimen0][r]{}} & 54.8\hspace{1pt}{\fontsize{7.2}{8}\selectfont\settowidth{\dimen0}{(\textminus{}43\%)}\makebox[\dimen0][r]{}} & 111.6\hspace{1pt}{\fontsize{7.2}{8}\selectfont\settowidth{\dimen0}{(\textminus{}28\%)}\makebox[\dimen0][r]{}} \\
\rowcolor{matcher2!10}
 & FP32 & 23.4 & 33.6 & 43.2 & 15.0\hspace{1pt}{\fontsize{7.2}{8}\selectfont\settowidth{\dimen0}{(\textminus{}31\%)}\makebox[\dimen0][r]{}} & 20.3\hspace{1pt}{\fontsize{7.2}{8}\selectfont\settowidth{\dimen0}{(\textminus{}31\%)}\makebox[\dimen0][r]{}} & 65.7\hspace{1pt}{\fontsize{7.2}{8}\selectfont\settowidth{\dimen0}{(\textminus{}43\%)}\makebox[\dimen0][r]{}} & 127.4\hspace{1pt}{\fontsize{7.2}{8}\selectfont\settowidth{\dimen0}{(\textminus{}28\%)}\makebox[\dimen0][r]{}} \\
\rowcolor{matcher2!10}
 & MP & 19.0 & 29.9 & 41.1 & 11.4\hspace{1pt}{\fontsize{7.2}{8}\selectfont\settowidth{\dimen0}{(\textminus{}31\%)}\makebox[\dimen0][r]{(\textminus{}24\%)}} & 14.3\hspace{1pt}{\fontsize{7.2}{8}\selectfont\settowidth{\dimen0}{(\textminus{}31\%)}\makebox[\dimen0][r]{(\textminus{}29\%)}} & 24.9\hspace{1pt}{\fontsize{7.2}{8}\selectfont\settowidth{\dimen0}{(\textminus{}43\%)}\makebox[\dimen0][r]{(\textminus{}62\%)}} & 116.1\hspace{1pt}{\fontsize{7.2}{8}\selectfont\settowidth{\dimen0}{(\textminus{}28\%)}\makebox[\dimen0][r]{(\textminus{}9\%)}} \\
\rowcolor{matcher2!10}
\multirow{-3}{*}{\textbf{SP+SG}~\cite{detone2018superpoint,sarlin2020superglue}} & FP16 & 20.2 & 33.0 & 43.6 & \textbf{9.1}\hspace{1pt}{\fontsize{7.2}{8}\selectfont\settowidth{\dimen0}{(\textminus{}31\%)}\makebox[\dimen0][r]{(\textminus{}39\%)}} & \textbf{11.9}\hspace{1pt}{\fontsize{7.2}{8}\selectfont\settowidth{\dimen0}{(\textminus{}31\%)}\makebox[\dimen0][r]{(\textminus{}41\%)}} & \textbf{20.9}\hspace{1pt}{\fontsize{7.2}{8}\selectfont\settowidth{\dimen0}{(\textminus{}43\%)}\makebox[\dimen0][r]{(\textminus{}68\%)}} & \textbf{98.7}\hspace{1pt}{\fontsize{7.2}{8}\selectfont\settowidth{\dimen0}{(\textminus{}28\%)}\makebox[\dimen0][r]{(\textminus{}23\%)}} \\
\rowcolor{matcher3!10}
 & FP32 & 17.9 & 29.0 & 40.5 & 12.2\hspace{1pt}{\fontsize{7.2}{8}\selectfont\settowidth{\dimen0}{(\textminus{}31\%)}\makebox[\dimen0][r]{}} & 18.3\hspace{1pt}{\fontsize{7.2}{8}\selectfont\settowidth{\dimen0}{(\textminus{}31\%)}\makebox[\dimen0][r]{}} & 57.2\hspace{1pt}{\fontsize{7.2}{8}\selectfont\settowidth{\dimen0}{(\textminus{}43\%)}\makebox[\dimen0][r]{}} & 113.3\hspace{1pt}{\fontsize{7.2}{8}\selectfont\settowidth{\dimen0}{(\textminus{}28\%)}\makebox[\dimen0][r]{}} \\
\rowcolor{matcher3!10}
 & MP & 18.6 & 30.4 & 42.8 & 9.0\hspace{1pt}{\fontsize{7.2}{8}\selectfont\settowidth{\dimen0}{(\textminus{}31\%)}\makebox[\dimen0][r]{(\textminus{}26\%)}} & 12.6\hspace{1pt}{\fontsize{7.2}{8}\selectfont\settowidth{\dimen0}{(\textminus{}31\%)}\makebox[\dimen0][r]{(\textminus{}31\%)}} & 23.0\hspace{1pt}{\fontsize{7.2}{8}\selectfont\settowidth{\dimen0}{(\textminus{}43\%)}\makebox[\dimen0][r]{(\textminus{}60\%)}} & 86.7\hspace{1pt}{\fontsize{7.2}{8}\selectfont\settowidth{\dimen0}{(\textminus{}28\%)}\makebox[\dimen0][r]{(\textminus{}23\%)}} \\
\rowcolor{matcher3!10}
\multirow{-3}{*}{\textbf{SP+LG}~\cite{detone2018superpoint,lindenberger2023lightglue}} & FP16 & 18.0 & 27.5 & 38.7 & \textbf{7.0}\hspace{1pt}{\fontsize{7.2}{8}\selectfont\settowidth{\dimen0}{(\textminus{}31\%)}\makebox[\dimen0][r]{(\textminus{}43\%)}} & \textbf{9.7}\hspace{1pt}{\fontsize{7.2}{8}\selectfont\settowidth{\dimen0}{(\textminus{}31\%)}\makebox[\dimen0][r]{(\textminus{}47\%)}} & \textbf{16.5}\hspace{1pt}{\fontsize{7.2}{8}\selectfont\settowidth{\dimen0}{(\textminus{}43\%)}\makebox[\dimen0][r]{(\textminus{}71\%)}} & \textbf{77.1}\hspace{1pt}{\fontsize{7.2}{8}\selectfont\settowidth{\dimen0}{(\textminus{}28\%)}\makebox[\dimen0][r]{(\textminus{}32\%)}} \\
\rowcolor{matcher4!10}
 & FP32 & 26.7 & 37.5 & 46.3 & 30.1\hspace{1pt}{\fontsize{7.2}{8}\selectfont\settowidth{\dimen0}{(\textminus{}31\%)}\makebox[\dimen0][r]{}} & 51.2\hspace{1pt}{\fontsize{7.2}{8}\selectfont\settowidth{\dimen0}{(\textminus{}31\%)}\makebox[\dimen0][r]{}} & 142.9\hspace{1pt}{\fontsize{7.2}{8}\selectfont\settowidth{\dimen0}{(\textminus{}43\%)}\makebox[\dimen0][r]{}} & 360.1\hspace{1pt}{\fontsize{7.2}{8}\selectfont\settowidth{\dimen0}{(\textminus{}28\%)}\makebox[\dimen0][r]{}} \\
\rowcolor{matcher4!10}
 & MP & 28.1 & 37.9 & 45.9 & 17.2\hspace{1pt}{\fontsize{7.2}{8}\selectfont\settowidth{\dimen0}{(\textminus{}31\%)}\makebox[\dimen0][r]{(\textminus{}43\%)}} & 30.0\hspace{1pt}{\fontsize{7.2}{8}\selectfont\settowidth{\dimen0}{(\textminus{}31\%)}\makebox[\dimen0][r]{(\textminus{}41\%)}} & 39.8\hspace{1pt}{\fontsize{7.2}{8}\selectfont\settowidth{\dimen0}{(\textminus{}43\%)}\makebox[\dimen0][r]{(\textminus{}72\%)}} & 171.2\hspace{1pt}{\fontsize{7.2}{8}\selectfont\settowidth{\dimen0}{(\textminus{}28\%)}\makebox[\dimen0][r]{(\textminus{}52\%)}} \\
\rowcolor{matcher4!10}
\multirow{-3}{*}{\textbf{LoFTR}~\cite{sun2021loftr}} & FP16 & 27.6 & 37.3 & 46.5 & \textbf{15.3}\hspace{1pt}{\fontsize{7.2}{8}\selectfont\settowidth{\dimen0}{(\textminus{}31\%)}\makebox[\dimen0][r]{(\textminus{}49\%)}} & \textbf{27.9}\hspace{1pt}{\fontsize{7.2}{8}\selectfont\settowidth{\dimen0}{(\textminus{}31\%)}\makebox[\dimen0][r]{(\textminus{}45\%)}} & \textbf{37.6}\hspace{1pt}{\fontsize{7.2}{8}\selectfont\settowidth{\dimen0}{(\textminus{}43\%)}\makebox[\dimen0][r]{(\textminus{}74\%)}} & \textbf{153.9}\hspace{1pt}{\fontsize{7.2}{8}\selectfont\settowidth{\dimen0}{(\textminus{}28\%)}\makebox[\dimen0][r]{(\textminus{}57\%)}} \\
\rowcolor{matcher5!10}
 & FP32 & 29.5 & 37.3 & 44.6 & 27.8\hspace{1pt}{\fontsize{7.2}{8}\selectfont\settowidth{\dimen0}{(\textminus{}31\%)}\makebox[\dimen0][r]{}} & 43.8\hspace{1pt}{\fontsize{7.2}{8}\selectfont\settowidth{\dimen0}{(\textminus{}31\%)}\makebox[\dimen0][r]{}} & 105.6\hspace{1pt}{\fontsize{7.2}{8}\selectfont\settowidth{\dimen0}{(\textminus{}43\%)}\makebox[\dimen0][r]{}} & 277.2\hspace{1pt}{\fontsize{7.2}{8}\selectfont\settowidth{\dimen0}{(\textminus{}28\%)}\makebox[\dimen0][r]{}} \\
\rowcolor{matcher5!10}
 & MP & 28.1 & 37.2 & 44.7 & 23.2\hspace{1pt}{\fontsize{7.2}{8}\selectfont\settowidth{\dimen0}{(\textminus{}31\%)}\makebox[\dimen0][r]{(\textminus{}17\%)}} & 35.8\hspace{1pt}{\fontsize{7.2}{8}\selectfont\settowidth{\dimen0}{(\textminus{}31\%)}\makebox[\dimen0][r]{(\textminus{}18\%)}} & 68.9\hspace{1pt}{\fontsize{7.2}{8}\selectfont\settowidth{\dimen0}{(\textminus{}43\%)}\makebox[\dimen0][r]{(\textminus{}35\%)}} & 218.0\hspace{1pt}{\fontsize{7.2}{8}\selectfont\settowidth{\dimen0}{(\textminus{}28\%)}\makebox[\dimen0][r]{(\textminus{}21\%)}} \\
\rowcolor{matcher5!10}
\multirow{-3}{*}{\textbf{ELoFTR}~\cite{wang2024eloftr}} & FP16 & 27.8 & 36.6 & 43.3 & \textbf{21.0}\hspace{1pt}{\fontsize{7.2}{8}\selectfont\settowidth{\dimen0}{(\textminus{}31\%)}\makebox[\dimen0][r]{(\textminus{}24\%)}} & \textbf{32.7}\hspace{1pt}{\fontsize{7.2}{8}\selectfont\settowidth{\dimen0}{(\textminus{}31\%)}\makebox[\dimen0][r]{(\textminus{}25\%)}} & \textbf{64.7}\hspace{1pt}{\fontsize{7.2}{8}\selectfont\settowidth{\dimen0}{(\textminus{}43\%)}\makebox[\dimen0][r]{(\textminus{}39\%)}} & \textbf{199.8}\hspace{1pt}{\fontsize{7.2}{8}\selectfont\settowidth{\dimen0}{(\textminus{}28\%)}\makebox[\dimen0][r]{(\textminus{}28\%)}} \\
\rowcolor{matcher6!12}
 & FP32 & 23.9 & 34.6 & 44.6 & 29.6\hspace{1pt}{\fontsize{7.2}{8}\selectfont\settowidth{\dimen0}{(\textminus{}31\%)}\makebox[\dimen0][r]{}} & 47.2\hspace{1pt}{\fontsize{7.2}{8}\selectfont\settowidth{\dimen0}{(\textminus{}31\%)}\makebox[\dimen0][r]{}} & 114.0\hspace{1pt}{\fontsize{7.2}{8}\selectfont\settowidth{\dimen0}{(\textminus{}43\%)}\makebox[\dimen0][r]{}} & 310.3\hspace{1pt}{\fontsize{7.2}{8}\selectfont\settowidth{\dimen0}{(\textminus{}28\%)}\makebox[\dimen0][r]{}} \\
\rowcolor{matcher6!12}
 & MP & 24.9 & 33.7 & 43.0 & 23.8\hspace{1pt}{\fontsize{7.2}{8}\selectfont\settowidth{\dimen0}{(\textminus{}31\%)}\makebox[\dimen0][r]{(\textminus{}20\%)}} & 35.9\hspace{1pt}{\fontsize{7.2}{8}\selectfont\settowidth{\dimen0}{(\textminus{}31\%)}\makebox[\dimen0][r]{(\textminus{}24\%)}} & 69.3\hspace{1pt}{\fontsize{7.2}{8}\selectfont\settowidth{\dimen0}{(\textminus{}43\%)}\makebox[\dimen0][r]{(\textminus{}39\%)}} & 226.3\hspace{1pt}{\fontsize{7.2}{8}\selectfont\settowidth{\dimen0}{(\textminus{}28\%)}\makebox[\dimen0][r]{(\textminus{}27\%)}} \\
\rowcolor{matcher6!12}
\multirow{-3}{*}{\textbf{MatchAnything}~\cite{He2025MatchAnything}} & FP16 & 22.4 & 34.2 & 45.0 & \textbf{22.1}\hspace{1pt}{\fontsize{7.2}{8}\selectfont\settowidth{\dimen0}{(\textminus{}31\%)}\makebox[\dimen0][r]{(\textminus{}25\%)}} & \textbf{34.0}\hspace{1pt}{\fontsize{7.2}{8}\selectfont\settowidth{\dimen0}{(\textminus{}31\%)}\makebox[\dimen0][r]{(\textminus{}28\%)}} & \textbf{67.1}\hspace{1pt}{\fontsize{7.2}{8}\selectfont\settowidth{\dimen0}{(\textminus{}43\%)}\makebox[\dimen0][r]{(\textminus{}41\%)}} & \textbf{207.0}\hspace{1pt}{\fontsize{7.2}{8}\selectfont\settowidth{\dimen0}{(\textminus{}28\%)}\makebox[\dimen0][r]{(\textminus{}33\%)}} \\
\rowcolor{matcher7!10}
 & FP32 & 47.2 & 58.8 & 67.8 & 651.9\hspace{1pt}{\fontsize{7.2}{8}\selectfont\settowidth{\dimen0}{(\textminus{}31\%)}\makebox[\dimen0][r]{}} & 1236.8\hspace{1pt}{\fontsize{7.2}{8}\selectfont\settowidth{\dimen0}{(\textminus{}31\%)}\makebox[\dimen0][r]{}} & 2649.4\hspace{1pt}{\fontsize{7.2}{8}\selectfont\settowidth{\dimen0}{(\textminus{}43\%)}\makebox[\dimen0][r]{}} & OOM\hspace{1pt}{\fontsize{7.2}{8}\selectfont\settowidth{\dimen0}{(\textminus{}28\%)}\makebox[\dimen0][r]{}} \\
\rowcolor{matcher7!10}
 & MP & 47.4 & 59.2 & 68.5 & \textbf{451.0}\hspace{1pt}{\fontsize{7.2}{8}\selectfont\settowidth{\dimen0}{(\textminus{}31\%)}\makebox[\dimen0][r]{(\textminus{}31\%)}} & \textbf{852.2}\hspace{1pt}{\fontsize{7.2}{8}\selectfont\settowidth{\dimen0}{(\textminus{}31\%)}\makebox[\dimen0][r]{(\textminus{}31\%)}} & \textbf{1517.1}\hspace{1pt}{\fontsize{7.2}{8}\selectfont\settowidth{\dimen0}{(\textminus{}43\%)}\makebox[\dimen0][r]{(\textminus{}43\%)}} & OOM\hspace{1pt}{\fontsize{7.2}{8}\selectfont\settowidth{\dimen0}{(\textminus{}28\%)}\makebox[\dimen0][r]{}} \\
\rowcolor{matcher7!10}
\multirow{-3}{*}{\textbf{DKM}~\cite{edstedt2023dkm}} & FP16 & \multicolumn{3}{c|}{Not supported} & \multicolumn{4}{c@{}}{Not supported} \\
\rowcolor{matcher8!10}
 & FP32 & 60.6 & 74.5 & 83.2 & 179.9\hspace{1pt}{\fontsize{7.2}{8}\selectfont\settowidth{\dimen0}{(\textminus{}31\%)}\makebox[\dimen0][r]{}} & 304.6\hspace{1pt}{\fontsize{7.2}{8}\selectfont\settowidth{\dimen0}{(\textminus{}31\%)}\makebox[\dimen0][r]{}} & 816.9\hspace{1pt}{\fontsize{7.2}{8}\selectfont\settowidth{\dimen0}{(\textminus{}43\%)}\makebox[\dimen0][r]{}} & 2388.7\hspace{1pt}{\fontsize{7.2}{8}\selectfont\settowidth{\dimen0}{(\textminus{}28\%)}\makebox[\dimen0][r]{}} \\
\rowcolor{matcher8!10}
 & MP & 60.1 & 74.8 & 83.8 & \textbf{148.7}\hspace{1pt}{\fontsize{7.2}{8}\selectfont\settowidth{\dimen0}{(\textminus{}31\%)}\makebox[\dimen0][r]{(\textminus{}17\%)}} & \textbf{243.7}\hspace{1pt}{\fontsize{7.2}{8}\selectfont\settowidth{\dimen0}{(\textminus{}31\%)}\makebox[\dimen0][r]{(\textminus{}20\%)}} & \textbf{559.7}\hspace{1pt}{\fontsize{7.2}{8}\selectfont\settowidth{\dimen0}{(\textminus{}43\%)}\makebox[\dimen0][r]{(\textminus{}31\%)}} & \textbf{1628.5}\hspace{1pt}{\fontsize{7.2}{8}\selectfont\settowidth{\dimen0}{(\textminus{}28\%)}\makebox[\dimen0][r]{(\textminus{}32\%)}} \\
\rowcolor{matcher8!10}
\multirow{-3}{*}{\textbf{MASt3R}~\cite{leroy2024mast3r}} & FP16 & \multicolumn{3}{c|}{Not supported} & \multicolumn{4}{c@{}}{Not supported} \\
\bottomrule
\end{tabularx}%
}}
\usebox{\precisiontablebox}
\par\vspace{4pt}
\begin{minipage}{\wd\precisiontablebox}
\footnotesize\raggedright
Parentheses: runtime change from FP32 (\%); bold: fastest precision per matcher/platform.\\
OOM: out of memory; Not supported: unsupported precision. Matcher colors follow Fig.~\ref{fig:accuracy-overview}.
\end{minipage}\endgroup

\end{table*}
\begin{figure*}[!t]
\centering
\begin{overpic}[width=\textwidth]{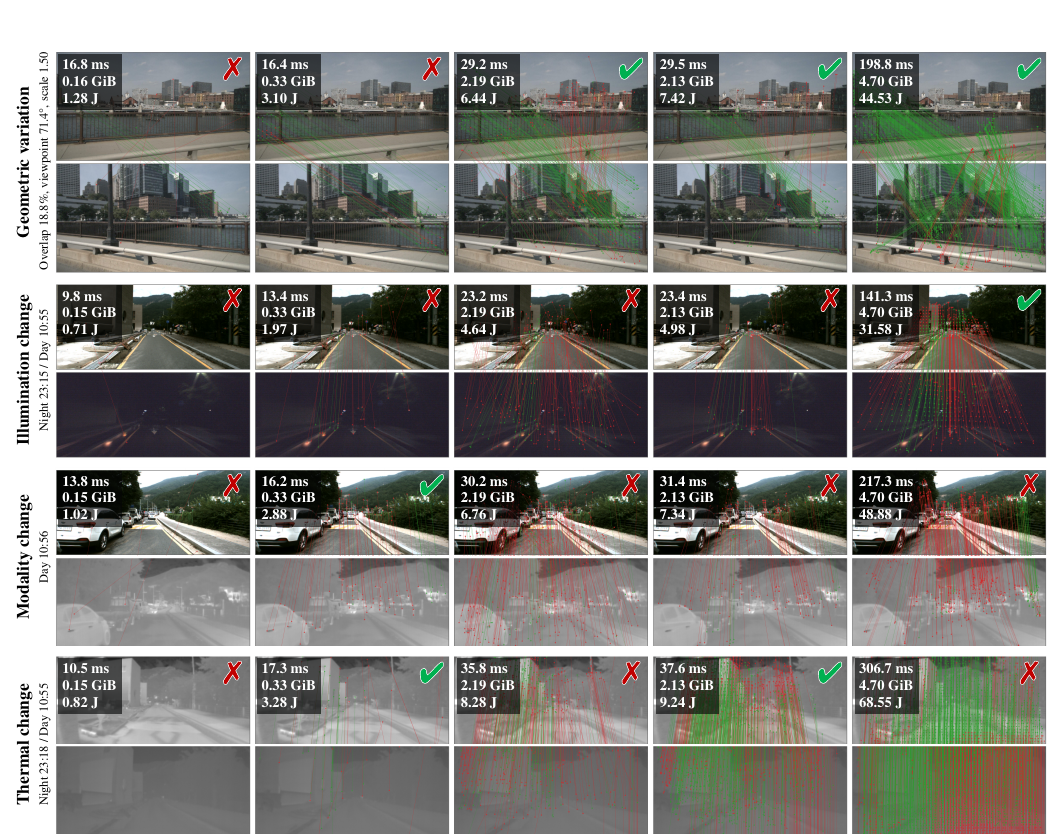}
\fontsize{7.5}{8}\selectfont
\put(14.55,76.37){\makebox(0,0){\textbf{SIFT (GPU)}~\cite{lowe2004sift}}}
\put(33.52,77.37){\makebox(0,0){\textbf{SuperPoint}~\cite{detone2018superpoint}+}}
\put(33.52,75.67){\makebox(0,0){\textbf{SuperGlue}~\cite{sarlin2020superglue}}}
\put(52.49,76.37){\makebox(0,0){\textbf{ELoFTR}~\cite{wang2024eloftr}}}
\put(71.46,76.37){\makebox(0,0){\textbf{MatchAnything}~\cite{He2025MatchAnything}}}
\put(90.43,76.37){\makebox(0,0){\textbf{MASt3R}~\cite{leroy2024mast3r}}}
\end{overpic}
\caption{Qualitative matching comparison at 512~px. Learned matchers use FP32; SIFT (GPU) uses Native. Rows follow Table~\ref{tab:visual-tasks}, with the same pair across matchers. Checks indicate pose errors within $10^\circ$; crosses indicate larger errors or failed estimates. Overlays show runtime (ms), peak GPU memory (GiB), and energy per pair (J), measured on RTX 5070 Ti. Green/red indicate correspondences geometrically consistent/inconsistent with the reference pose. Thermal images and correspondence overlays are vertically center-cropped for display only.}
\label{fig:qualitative-matching}
\end{figure*}

\section{Results}
\label{sec:results}
We first compare matching accuracy across visual conditions, then examine hardware-dependent runtime, resource costs, and precision savings.

\subsection{Accuracy across Visual Conditions and Resolutions}
\label{sec:results_accuracy}

Fig.~\ref{fig:accuracy-overview}(a) compares pose AUC@10$^\circ$ across four visual conditions and image resolutions on RTX 5070 Ti, using FP32 for learned matchers and Native for ORB (CUDA) and SIFT (GPU). On \texttt{RUBIK}, MASt3R reaches a pose AUC of 74.5\% at 512~px. Increasing resolution from 512 to 1024~px improves pose AUC for ELoFTR, MatchAnything, and DKM, while MASt3R declines to 28.8\%. The resolution yielding the highest accuracy therefore depends on the matcher.

Day--night visible matching remains challenging for all evaluated matchers: at 512~px, none exceeds 10\% pose AUC@10$^\circ$. For thermal day--night matching, SuperPoint+SuperGlue achieves 23.9\%, followed by DKM at 20.1\%, while MASt3R reaches 3.1\% despite its strong performance on \texttt{RUBIK}.

For daytime visible--thermal matching, increasing resolution produces different trends across matchers. ELoFTR's pose AUC@10$^\circ$ decreases from 26.6\% at 512~px to 6.7\% at 1024~px. LoFTR and MASt3R also decline, whereas MatchAnything improves from 10.1\% to 22.9\%. These results support selecting resolution according to both matcher and visual condition.

\subsection{Accuracy--Runtime Trade-offs across Hardware}
\label{sec:results_hardware_tradeoffs}
Fig.~\ref{fig:accuracy-overview}(b) compares accuracy--runtime trade-offs across hardware platforms on \texttt{RUBIK} at 256 and 512~px, using FP32 for learned matchers and Native for ORB (CUDA) and SIFT (GPU). Accuracy is evaluated on RTX 5070 Ti; each platform's runtime places that accuracy at a different computational cost. The plot thus shows which matcher and resolution combinations combine the desired accuracy with the target platform's runtime budget.

\subsection{Computational Cost across Hardware Platforms}
\label{sec:results_computational_cost}

Table~\ref{tab:hardware-cost-comparison} compares computational costs on \texttt{RUBIK} at 512~px, using FP32 for learned matchers and Native for ORB (CUDA) and SIFT (GPU). For the same SuperPoint+LightGlue configuration, median runtime increases from 12.2 and 18.3~ms on RTX 5070 Ti and \mbox{RTX 5060 Ti} to 57.2~ms on Thor and 113.3~ms on Orin Nano. Thus, a configuration that meets a runtime budget on a workstation can exceed it on an onboard platform.

Table~\ref{tab:hardware-cost-comparison} reports memory and energy costs relative to SuperPoint+SuperGlue on each platform. On RTX 5070 Ti, SuperPoint+SuperGlue uses 0.330~GiB and 3.00~J per pair, whereas DKM uses 13.44~GiB and 163.46~J, corresponding to 40.7 and 54.4 times these costs.

Memory and energy requirements also vary substantially with input resolution. In the FP32 evaluation on \texttt{RUBIK}, increasing resolution from 512 to 1024~px raises ELoFTR's GPU memory on RTX 5070 Ti from 2.19 to 10.24~GiB. Under the same conditions, MASt3R's energy per pair increases from 44.21 to 330.60~J. These cost changes can be compared with the resolution-dependent accuracy in Section~\ref{sec:results_accuracy} to inform the choice of input resolution.

\subsection{Resource Savings through Reduced Precision}
\label{sec:results_precision}
Table~\ref{tab:precision-accuracy-runtime} examines whether reduced precision can bring a
configuration within a deployment budget while retaining matching accuracy. It compares FP32,
MP, and FP16 at a fixed 512~px resolution on \texttt{RUBIK}.

On \texttt{RUBIK} at 512~px, switching SuperPoint+SuperGlue from FP32 to FP16 reduces runtime by 68\%, GPU memory by 65\%, and energy per pair by 73\% on Thor. Under the same conditions, LoFTR's memory changes by less than 1\%, while its energy decreases by 78\%, showing that the reductions across metrics depend on the matcher.

For example, SuperPoint+SuperGlue on Thor falls from 65.7~ms in FP32 to 20.9~ms in FP16,
meeting a 50~ms budget that FP32 exceeds.

Table~\ref{tab:precision-accuracy-runtime} reports pose AUC at 5$^\circ$, 10$^\circ$, and 20$^\circ$ for each evaluated precision. On \texttt{RUBIK} at 512~px, the largest absolute change from FP32 across these thresholds is 4.42 percentage points, observed for SuperPoint+SuperGlue in MP at 5$^\circ$. These accuracy changes must be weighed against the runtime savings when selecting a precision mode.

Precision also changes which configurations can run under memory constraints. On \texttt{RUBIK} at 1024~px on Orin Nano, ELoFTR and MatchAnything fail with out-of-memory errors in FP32 and MP, but run in FP16 with peak CUDA allocations of 4.29 and 4.33~GiB, respectively. Finding configurations that can run on the target platform therefore requires considering precision alongside the matcher and resolution.

\subsection{Qualitative Matching Results}
\label{sec:results_qualitative}

Fig.~\ref{fig:qualitative-matching} shows viewpoint variation in \texttt{RUBIK} and day--night and visible--thermal image pairs from \texttt{STheReO} together with the correspondences returned by each matcher. These examples illustrate how correspondence distributions and pose estimation outcomes differ across matchers for the same image pair. SIFT (GPU) returns few or no correspondences in the displayed day--night examples, while SuperPoint+SuperGlue estimates a pose with an error within 10$^\circ$ in the daytime visible--thermal example.

\subsection{Deployment Examples}
\label{sec:results_selection}

\begin{figure}[!t]
\centering
\includegraphics[width=\columnwidth]{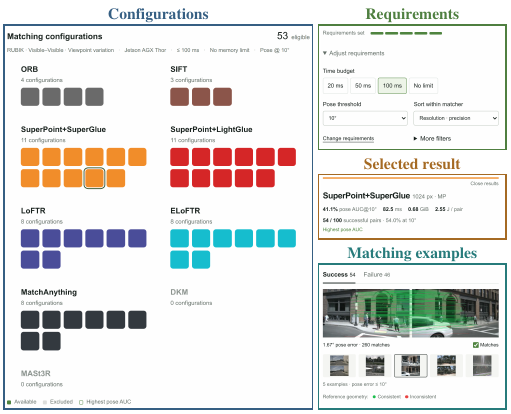}
\caption{\matchercompass{} Web GUI on RUBIK with Jetson AGX Thor and a 100\,ms runtime budget. Tiles represent feasible configurations; selecting a tile reveals its accuracy and resource costs.}
\label{fig:web-selection-pipeline}
\end{figure}

Fig.~\ref{fig:web-gui-demo} connects user requirements to measured candidates; the GUI is shown in Fig.~\ref{fig:web-selection-pipeline}. For example, on \texttt{RUBIK} at 512~px, a 20~ms budget on Thor admits ORB (CUDA) in Native and SP+LG in FP16 (Table~\ref{tab:precision-accuracy-runtime}). Users compare the returned candidates by accuracy and resource cost.

\section{Findings and Deployment Guidelines}
\label{sec:discussion}

\textbf{Hardware-specific budget choices.}
On \texttt{RUBIK} at 512~px, Orin Nano's learned candidates within 100~ms are SP+LG
in MP/FP16 and SP+SG in FP16. SP+LG FP16 offers more timing margin at 77.1~ms;
SP+SG FP16 provides higher pose AUC@10$^\circ$ (33.0\% versus 27.5\%) at 98.7~ms.
On Thor, SP+LG FP16 also meets 20~ms. These medians cover matching calls, not the full processing cycle.

\textbf{Practical benefits of reduced precision.}
On \texttt{RUBIK} at 512~px, MP/FP16 reduce runtime with pose AUC changes of at most
4.42 percentage points from FP32. On Thor under these conditions, switching SP+SG from FP32
to FP16 reduces runtime, measured peak GPU memory, and energy by 68\%, 65\%, and 73\%,
respectively. Under the same conditions, LoFTR's peak GPU memory changes by less than 1\%.
FP16 merits consideration, with task-specific accuracy and resource checks.

\begin{figure}[!t]
\centering
\includegraphics[width=\columnwidth]{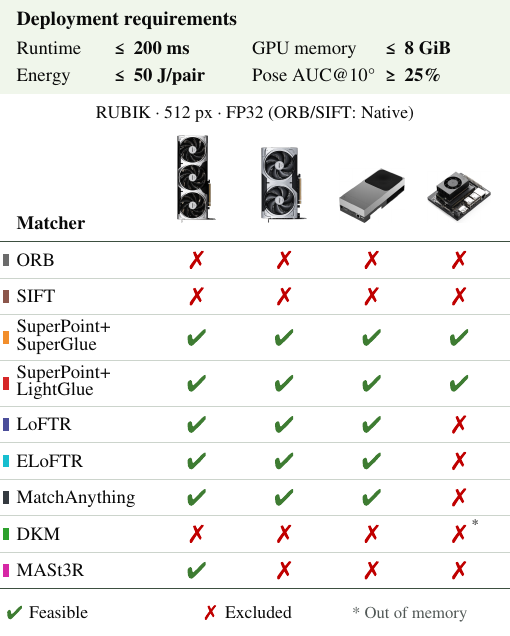}
\caption{Matcher feasibility under the shown requirements. Representative hardware images, left to right: RTX 5070 Ti, RTX 5060 Ti, Jetson AGX Thor, and Jetson Orin Nano.}
\label{fig:deployment-guide}
\end{figure}

\textbf{Deployment costs beyond runtime.}
On RTX 5070 Ti, DKM uses 40.7 times the GPU memory and 54.4 times the energy of
SP+SG. On Orin Nano at 1024~px, ELoFTR and MatchAnything fail in FP32/MP because
of insufficient GPU memory (OOM), but run in FP16 with peaks of 4.29 and 4.33~GiB.
Memory feasibility and energy must therefore be checked alongside latency (Fig.~\ref{fig:deployment-guide}).

\textbf{Different rankings for cross-modal matching.}
MASt3R leads on \texttt{RUBIK} at 512~px but not on visible--thermal pairs
(Fig.~\ref{fig:accuracy-overview}(a)). For visible--thermal matching, ELoFTR exceeds
MatchAnything in pose AUC@10$^\circ$ at 512~px; at 1024~px, MatchAnything leads with
pose AUC@10$^\circ$ of 22.9\% versus 6.7\%.
RGB rankings therefore do not transfer directly: select the matcher and resolution
jointly on the target modality pair.

\textbf{Uncertain returns from higher resolution.}
At 1024 versus 512~px, \texttt{RUBIK} accuracy improves for ELoFTR, MatchAnything, and DKM but declines for MASt3R. On RTX 5070 Ti, ELoFTR memory rises from
2.19 to 10.24~GiB and MASt3R energy from 44.21 to 330.60~J; ELoFTR is slower than LoFTR
at 128~px but faster at 1024~px. Choose resolution for measured accuracy and cost.

\textbf{Geometric validity over correspondence count.}
On visible--thermal pairs, ELoFTR in FP32 returns more correspondences at 1024 than at 512~px,
yet pose AUC@10$^\circ$ \mbox{decreases}. Pose estimates are returned for all 100 pairs at both resolutions.
Neither correspondence count nor successful pose recovery alone establishes matching quality;
selection should be based on the resulting pose errors.

\section{Conclusion}
\label{sec:conclusion}

We present \matchercompass{} to support local feature matcher selection for the visual conditions and computing hardware of field robots. We evaluated nine matching pipelines across four visual conditions, four input resolutions, and supported numerical precisions, linking pose accuracy to runtime, GPU memory, and energy per pair on four hardware platforms. The results show that increasing resolution does not always improve accuracy, and that hardware and precision change which configurations satisfy time and memory constraints. \matchercompass{} provides a measurement-based selection guide for comparing all configurations that meet these constraints in terms of accuracy and resource cost.

\bibliographystyle{ieeetr}
{\footnotesize
\bibliography{references}
}

\end{document}